\documentclass{article}

\usepackage[preprint]{neurips_2024}

\usepackage[utf8]{inputenc} % allow utf-8 input
\usepackage[T1]{fontenc}    % use 8-bit T1 fonts
\usepackage{hyperref}       % hyperlinks
\usepackage{url}            % simple URL typesetting
\usepackage{booktabs}       % professional-quality tables
\usepackage{amsfonts}       % blackboard math symbols
\usepackage{nicefrac}       % compact symbols for 1/2, etc.
\usepackage{microtype}      % microtypography
\usepackage{xcolor}         % colors
\usepackage{graphicx}
\usepackage{multirow}
\usepackage{subcaption}

\usepackage{url}
\usepackage{hyperref}

\title{Improving the Classification Effect of Clinical Images of Diseases for Multi-Source Privacy Protection}

\author{%
  Bowen Tian\thanks{ These authors have the same contributions }\\
  The Hong Kong University of Science and Technology (Guangzhou)\\
    Deep Interdisciplinary Intelligence Lab ($DI^2 Lab$)\\
  China University of Mining and Technology(Beijing)\\
  \texttt{bowentian@hkust-gz.edu.cn} \\
  \And
  Zhengyang Xu\footnotemark[1]\\
  Department of Artificial Intelligence\\
  China University of Mining and Technology(Beijing)\\
  \texttt{x3542049703@163.com} \\
  \And
  Zihao Yin\\
  The Hong Kong University of Science and Technology (Guangzhou)\\
  Financial Technology Thrust\\
  The Chinese University of Hong Kong\\
  \texttt{zihaoyin81@gmail.com} \\
  \And
  Jingying Wang\\
  Department of Artificial Intelligence\\
  China University of Mining and Technology(Beijing)\\
  \texttt{jingyingwang2002@163.com} \\
  \And
Yutao Yue\thanks{Correspondence to Yutao Yue \{yutaoyue@hkust-gz.edu.cn\}}. \\
  The Hong Kong University of Science and Technology (Guangzhou) \\
  Institute of Deep Perception Technology, JITRI\\
  Deep Interdisciplinary Intelligence Lab ($DI^2 Lab$)\\
  \texttt{yutaoyue@hkust-gz.edu.cn}\\
}

\begin{document}

\maketitle

\begin{abstract}
Privacy data protection in the medical field poses challenges to data sharing, limiting the ability to integrate data across hospitals for training high-precision auxiliary diagnostic models. Traditional centralized training methods are difficult to apply due to violations of privacy protection principles. Federated learning, as a distributed machine learning framework, helps address this issue, but it requires multiple hospitals to participate in training simultaneously, which is hard to achieve in practice. To address these challenges, we propose a medical privacy data training framework based on data vectors. This framework allows each hospital to fine-tune pre-trained models on private data, calculate data vectors (representing the optimization direction of model parameters in the solution space), and sum them up to generate synthetic weights that integrate model information from multiple hospitals. This approach enhances model performance without exchanging private data or requiring synchronous training. Experimental results demonstrate that this method effectively utilizes dispersed private data resources while protecting patient privacy. The auxiliary diagnostic model trained using this approach significantly outperforms models trained independently by a single hospital, providing a new perspective for resolving the conflict between medical data privacy protection and model training and advancing the development of medical intelligence. 
\end{abstract}

\section{Introduction}

In the medical field \cite{lai2022predicting}, the protection of private data is crucial, but it also poses challenges for data sharing \cite{khalid2023privacy}. Due to privacy regulations, it is difficult for hospitals to directly share sensitive diagnostic data, making it extremely challenging to integrate private data from multiple hospitals to train high-accuracy auxiliary diagnostic models. Traditional centralized training methods \cite{cai2020review} can not be applied in this scenario because they require data to be shared across institutions and processed centrally, which violates the principles of privacy protection.

Federated Learning (FL)\cite{wen2023survey}, as a distributed machine learning framework, has been proposed to address this issue. It allows multiple participants to jointly train models without sharing data. However, the application of FL also faces challenges. Among them, the model parameter averaging method often fails to achieve good results due to factors such as permutation invariance in deep networks and linear connection barriers, which hinder the full utilization of data from each client. Although gradient aggregation-based methods have improved performance, they require each client to participate in the same training process during the same time period to ensure the effectiveness of training. This stringent requirement is rarely used in practice.

In recent years, direct editing of model parameters has gradually attracted the attention of researchers, with many works exploring the distribution and morphology of model parameter spaces. Recent studies such as \cite{frankle2020linear,wortsman2022robust,matena2022merging,wortsman2022model,lai2023faithful,li2022branch,ainsworth2022git,don2022cold,ilharco2022editing} have demonstrated that despite the nonlinear activation functions in deep learning models \cite{lai2023multimodal,lai2024ftsframeworkfaithfultimesieve}, linear interpolation in the model parameter space can also directionally alter the model's capabilities to some extent. This suggests that the information contained in the parameters may be directly applicable, and this technique has shown promising results in many fields.

Inspired by the above works, we propose a novel training framework for medical privacy data based on data vectors to overcome these challenges in the field of medical auxiliary diagnostics. We first assume that each hospital client uses the same pre-trained model \cite{han2021pre} architecture as the based model. They fine-tune an auxiliary diagnostic model for the same disease on their respective private data. The data vector is obtained by calculating the difference between the fine-tuned weights and the pre-trained weights at each hospital client. This data vector represents the direction of model parameter movement in the solution space under the fine-tuning of local data at that client \cite{goyal2023finetune}. Since each hospital's local data is different, the direction of optimization also differs. Based on the assumption that the data vectors generated by fine-tuning at different hospitals are complementary, we propose to add the data vectors calculated by each hospital's independently fine-tuned auxiliary diagnostic model and then add the synthesized data vector to the pre-trained weights to generate synthesized weights \cite{ilharco2022editing}. These synthesized weights can be considered as integrating model information from multiple hospitals to obtain a more accurate optimization direction, thereby improving model's performance. This method does not require the exchange of privacy data or simultaneous training by all hospitals, greatly reducing the barriers to practical application.

Through experimental testing on three different types of publicly available medical classification datasets, we found that this data vector-based hybrid model significantly outperforms any model trained independently by a single hospital. These results indicate that our method can effectively utilize dispersed private data resources from various hospitals to train higher-accuracy auxiliary diagnostic models while protecting patient privacy. It not only provides a new approach to resolving the conflict between medical data privacy protection and model training but also contributes to advancing the development of medical intelligence.

In summary, the contributions of our paper are as follows:

\begin{itemize}
    \item[1)] We are the first to integrate models trained on multiple small datasets through a model parameter mixing approach, obtaining a hybrid model with improved performance without the need for additional training. This technique holds great promise for application in the field of deep learning-based medical auxiliary diagnosis, where privacy concerns are significant.
    \item[2)] We conducted experimental tests of our method on three completely different types of publicly available medical datasets. Comparative analysis demonstrates that this method exhibits excellent adaptability across various medical domains and significantly outperforms general model parameter averaging methods.
    \item[3)] We provide a theoretical explanation for model mixing to some extent, preliminarily elucidating the rationale behind the enhanced performance of hybrid models derived from models trained on different datasets. Simultaneously, we propose potential directions for future exploration in model parameter mixing.
\end{itemize}

\section{Problem Setting}
\label{problem_setting}
To fit the assumptions of the use case, we abstract the data owned by different hospitals into multiple training sets $\mathbf{D} = \{\mathbf{D}_i|i \in (0,N]\}$, Each training set can be represented as $\mathbf{D}_i = \{(x_j,y_j)|j\in N_i\}$, these datasets is about the same disease, where $x_j$ and $y_j$ represent the clinical images in the dataset and their corresponding diagnostic labels, respectively. Suppose the selected pre-trained model has a weight $\theta_{pre}$, The pre-trained weights were used to independently train on the above $N$ datasets to obtain $N$ downstream fine-tuning models. The fine-tuned weights can be expressed as $\{\theta_{i}|i\in (0,N]\}$, where $\theta_i$ represents the model parameters obtained by fine-tuning on $\mathbf{D}_i$. Our goal is to use these independently trained weights to synthesize a model that performs better without accessing the opponent's training set and without having to retrain at the same time.

\section{Method}

\label{method}
\begin{figure*}[t]
  \includegraphics[width=\linewidth]{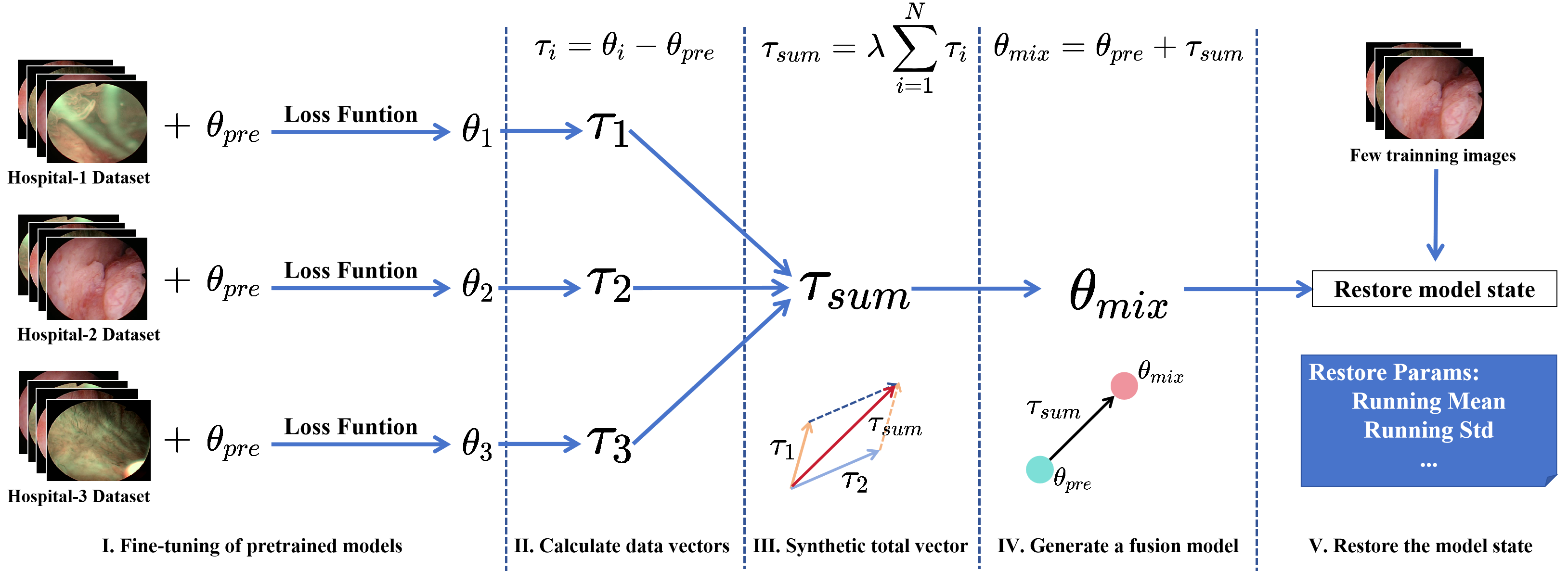}
  \caption{This is the overview of our proposed methodology, and each stage is described in detail in \autoref{method}.}
  \label{fig:overview}
\end{figure*}

This section details how our proposed data vector approach can be applied to the blending of medically assisted diagnostic models, as shown in \autoref{fig:overview}, and the overall approach is divided into five parts, each of which is described in detail below.

\subsection{Fine-tuning of pretrained models}

In the field of medical image classification, fine-tuning pre-trained models for downstream tasks generally achieves faster convergence and superior performance compared to training models from scratch. Indeed, most existing methods in this domain are implemented based on fine-tuning pre-trained models. To better align with practical application scenarios, we stipulate that all models are based on a unified model architecture (ResNet50 \cite{he2016deep} is chosen in this paper) and the same pre-trained weights (obtained from pre-training on the ImageNet \cite{deng2009imagenet} dataset).

For the $N$ datasets mentioned in the \autoref{problem_setting}, we uniformly adopt cross-entropy loss as the loss function during the fine-tuning stage. The fine-tuning process can be succinctly described as follows:
$$\theta_{i} = argmin_{\theta}\{\sum_{j=1}^{N_i}\mathcal{H}(p(x_i|\theta),y_j)\}$$
where $\mathcal{H}$ denotes the cross-entropy loss function.

\subsection{Calculate data vectors}

The data vector is used to capture the direction of movement of model weights within the solution space relative to the pre-trained model weights when fine-tuning is performed on a specific dataset. Specifically, it aims to reveal the potential direction of movement that the model weights may take to reach the optimal solution given a particular dataset.

Since the weights of each layer in deep learning can be represented as a vector in a high-dimensional space, to accurately describe the subsequent computational process, the weights of the model can be uniformly represented by layers as follows:
\begin{equation}
\theta = \{w^i \in \mathbb{R}^{d_i} |i \in (0,M]\}
\end{equation}
where $w^i$ represents the weight of layer $i$, $d_i$ represents the dimension of layer $i$, and $M$ represents the total number of layers of the model.

Based on the above definition of model weights, the calculation of the data vector can be expressed as:
$$\tau_j = \theta_j - \theta_{pre} = \{(w_j^i-w_{pre}^i) \in \mathbb{R}^{d_i} |i \in (0,M]\}$$
The computed data vectors have the same layer structure as the pre-trained model, and the dimensions of each layer are the same.

\subsection{Synthetic total vector}

Given the inherent randomness in model weight optimization during fine-tuning and the inevitable presence of noise in each dataset, this often leads to varying degrees of noise accompanying the direction of model weight optimization when trained on a single dataset. Based on prior works \cite{frankle2020linear,matena2022merging,ilharco2022editing}, the feasibility of interpolation has been demonstrated at the model weight level. Therefore, we proceed from an assumption: the weights of pre-trained models after fine-tuning are located within the same flat loss valley \cite{li2023deep}, allowing for the linear addition of multiple data vectors. Since the data vectors calculated from models fine-tuned on different datasets for the same task all exhibit a trend towards moving in the direction of correct weights, their sum can, to some extent, offset the noise in the parameter movement path and enhance the trend towards the correct weight direction, thereby obtaining a more accurate and reliable direction of weight movement.

The addition between any two of these data vectors can be defined as:
\begin{equation}
\tau_n + \tau_m = \{(w^i_n + w^i_m)\in \mathbb{R}^{d_i}|i\in (0,M]\}
\end{equation}

Given that all data vectors share the same layer structure and dimensions, the linear addition between data vectors can be succinctly represented as:
$$\tau_{sum} = \lambda \sum_{i=1}^N \tau_i$$
where $\tau_i$ represents the data vector of the model weights fine-tuned on $\mathbf{D}_i$ relative to the pre-trained weights $\theta_{pre}$. $\tau_{sum}$, on the other hand, represents the result of range-controlled (based on hyperparameter $\lambda$) data vectors obtained from the sum of data vectors generated by different dataset.

\subsection{Generate a fusion model}

In the above summary, we have introduced the process of obtaining the data vector and the way to merge the data vector, naturally, we need to apply the merged data vector to the original pre-trained model to obtain the complete usable fusion model, this step can be simply expressed as:
\begin{equation}
\theta_{mix} = \theta_{pre} + \tau_{sum} = \{(w^i_{pre} + w^i_{sum})\in \mathbb{R}^{d_i}|i\in (0,M]\}
\end{equation}
where $\theta_{mix}$ represents the weights of the fusion model.
\subsection{Restore the model state}

Directly using the merged model often fails to achieve the desired good performance, because some parameters of the model are closely related to the training state, such as the \textit{RunningMean} and \textit{RunningStd} parameters in the commonly used batch normalization layer \cite{ioffe2015batch} in image processing models. During the process of generating data vectors, these parameters are inevitably affected, leading to deviations in model predictions. Therefore, we need to adjust the hybrid model using a small amount of data samples to restore it to its normal state. There are various methods to achieve this step, and the specific method adopted in this work will be elaborated in the experimental section.

\section{Experiments}
\label{experiments}

\begin{table}[htbp]
\centering
\caption{Retina}\label{tab:retina}
\begin{tabular}{cccccc}
\hline
Part Index & Params Mean\cite{wortsman2022model} & Data Vector(Ours) & Random Vector & Base Model & Full Training \\
\hline
1 & 51.38 & 70.71 & 30.14 & 57.45 & \multirow{4}{*}{73.48} \\
% \cline{1,5} \\
2 & 49.72 & 69.61 & 15.48 & 60.77 & ~ \\
3 & 50.83 & 72.37 & 28.90 & 65.19 & ~ \\
4 & 53.03 & 71.82 & 39.92 & 64.64 & ~ \\
\hline
\end{tabular}
\end{table}

\begin{table}[htbp]
\centering
\caption{Endoscopic Bladder Tissue}\label{tab:EBT}
\begin{tabular}{cccccc}
\hline
Part Index & Params Mean\cite{wortsman2022model} & Data Vector(Ours) & Random Vector & Base Model & Full Training \\
\hline
1 & 62.43 & 65.07 & 20.1 & 57.67 & \multirow{4}{*}{63.49} \\
% \cline{1,5} \\
2 & 60.31 & 68.25 & 13.22 & 60.31 & ~ \\
3 & 60.84 & 66.67 & 28.04 & 58.73 & ~ \\
4 & 61.37 & 65.07 & 27.51 & 54.49 & ~ \\
\hline
\end{tabular}
\end{table}

\begin{table}[htbp]
\centering
\caption{PAD-UFES-20}\label{tab:skin}
\begin{tabular}{cccccc}
\hline
Part Index & Params Mean\cite{wortsman2022model} & Data Vector(Ours) & Random Vector & Base Model & Full Training \\
\hline
1 & 58.59 & 67.57 & 31.25 & 63.28 & \multirow{4}{*}{68.36}\\
2 & 60.93 & 66.01 & 5.85 & 64.45 & ~ \\
3 & 57.42 & 66.4 & 10.54 & 67.57 & ~ \\
4 & 60.93 & 66.79 & 21.48 & 63.67 & ~ \\
\hline
\end{tabular}
\end{table}

To comprehensively evaluate our proposed method, we carefully selected three medical diagnosis datasets covering different disease types: the PAD-UFES-20 \cite{pacheco2020pad} dataset for skin disease diagnosis, the Retina \cite{Retina} dataset for retinal fundus image classification, and the Endoscopic Bladder Tissue \cite{lazo2023semi} dataset for endoscopic image classification of bladder tissues. Detailed information about these datasets will be listed in \autoref{appendix}.

\subsection{Implementation details}

we chose ResNet50 as the classification network and utilized its pre-trained weights on ImageNet as the unified default weights. Given the difference in the number of classes between the classification head of the pre-trained ResNet50 model and our selected downstream task classification heads, we added a new classification layer after the output layer of ResNet50 to ensure its compatibility with the classification requirements of downstream tasks.

For each dataset, we equally divided it into four parts to simulate the private data environment held by different hospitals. Subsequently, we fine-tuned the model on these four different datasets separately. The parameter settings used during fine-tuning are described below:
\begin{table}[h]
\centering
\caption{Parameter settings}\label{tab:setting}
\begin{tabular}{c|c}
\hline
Learning Rate & 1e-5 \\
\hline
Epochs & 64 \\
\hline
Batch Size & 32 \\
\hline
Optimizer & Adam \\
\hline
Loss Func & Weighted Cross Entropy Loss \\
\hline
Device & NVIDIA 3090 \\ 
\hline
Random Seed & 3315 \\

\hline
\end{tabular}
\end{table}

Through fine-tuning, we obtained four models with the same architecture but different parameters. To further enhance model performance, we mixed the parameters of these four models to generate a more powerful ensemble model. Since the classification layer we added did not contain pre-trained parameters, we only considered layers with pre-trained parameters and ignored the classification layer when using data vectors. After calculating the corresponding data vectors for the four models, we adopted the method proposed in Section 3 to synthesize these data vectors into a unified data vector, with a coefficient lambda of 0.5 uniformly selected. After calculating the synthesized data vector, we could directly apply it to the pre-trained parameters to form mixed parameters. For the classification head, we simply averaged the classification head parameters of the four models to obtain a more stable classification effect.

After completing parameter mixing, we obtained the final ensemble model. Next, we trained the ensemble model on the four datasets for one epoch to restore the model state. This approach not only complied with privacy agreements between hospitals but also had extremely low training costs.

\subsection{Interpretation of results}

In our experimental results, we compared four different methods and provided results from fine-tuning using the complete dataset as a reference. Among them, "Params Mean \cite{wortsman2022model}" represents the average model obtained by directly averaging all parameters of the four models, which is the most commonly used method in previous parameter mixing, and we used it as the main comparison method; "Data Vector" represents the results obtained using the method proposed in this paper; "Random Vector" uses randomly generated vectors instead of Data Vectors to prove the effectiveness of Data Vectors; "Base Model" represents the effect of directly fine-tuning the original pre-trained model on the four datasets; and "Full Training" represents the results from fine-tuning using the complete dataset. All values in the experiment results table are accuracy metrics.

As observed from \autoref{tab:retina} \autoref{tab:EBT} \autoref{tab:skin}, the method based on Data Vector significantly outperforms direct fine-tuning on any dataset fragment in terms of performance and achieves results close to or even surpassing those of fine-tuning using the complete dataset. Taking the Endoscopic Bladder Tissue dataset as an example, the highest accuracy achieved by models fine-tuned on partial datasets is only 60.31\%, while our Data Vector method can directly achieve an accuracy of 68.25\%. In comparison, the best accuracy achieved by the method of directly averaging parameters is only 62.43\%, which strikingly highlights the superiority of the Data Vector method. Even when fine-tuning is performed using the complete dataset, the best accuracy achievable by the model is only 63.49\%, which is still significantly lower than the performance of the hybrid model generated by the Data Vector method. Furthermore, when randomly generated vectors are used instead of Data Vector, the best accuracy is only 28.04\%, indicating that Data Vector not only effectively indicates the correct direction of parameter movement for model optimization but also that its effectiveness is not accidental or merely due to randomness enabling the model to escape from local minima regions.

\begin{figure}
    \centering
    \begin{subfigure}{0.47\linewidth}
        \centering
        \includegraphics[width=\linewidth]{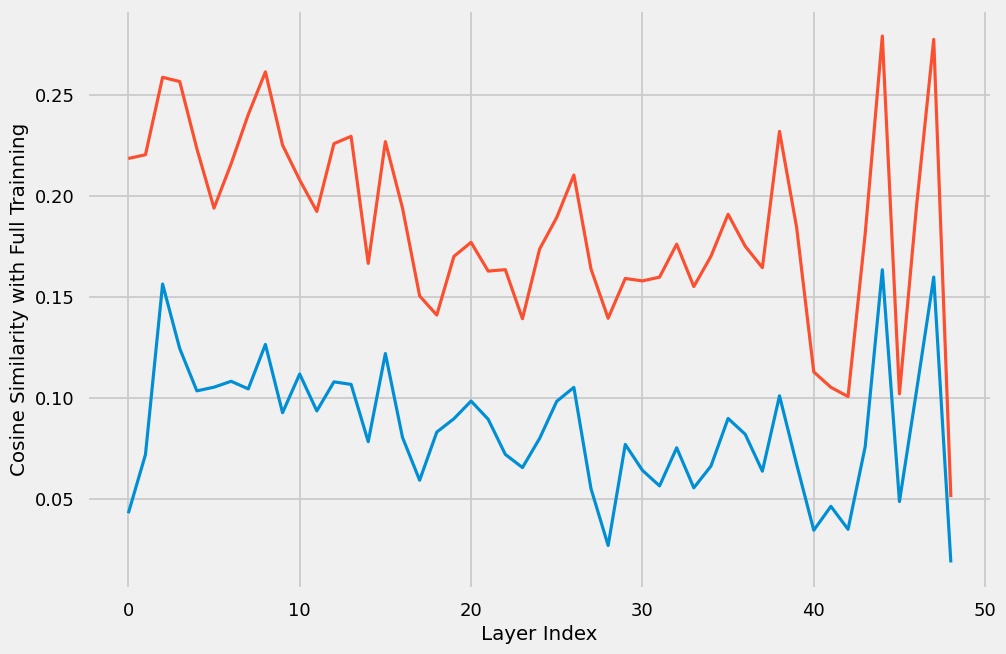}
        \caption{part-1}
        \label{fig:part1}
    \end{subfigure}
    \begin{subfigure}{0.47\linewidth}
        \centering
        \includegraphics[width=\linewidth]{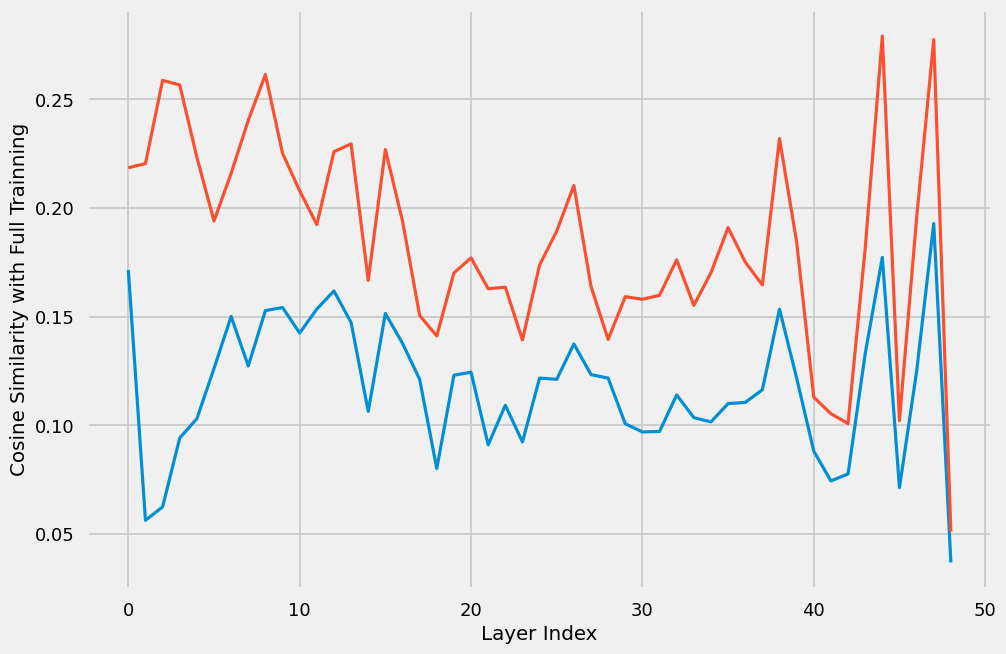}
        \caption{part-2}
        \label{fig:part2}
    \end{subfigure} \\
    \begin{subfigure}{0.47\linewidth}
        \centering
        \includegraphics[width=\linewidth]{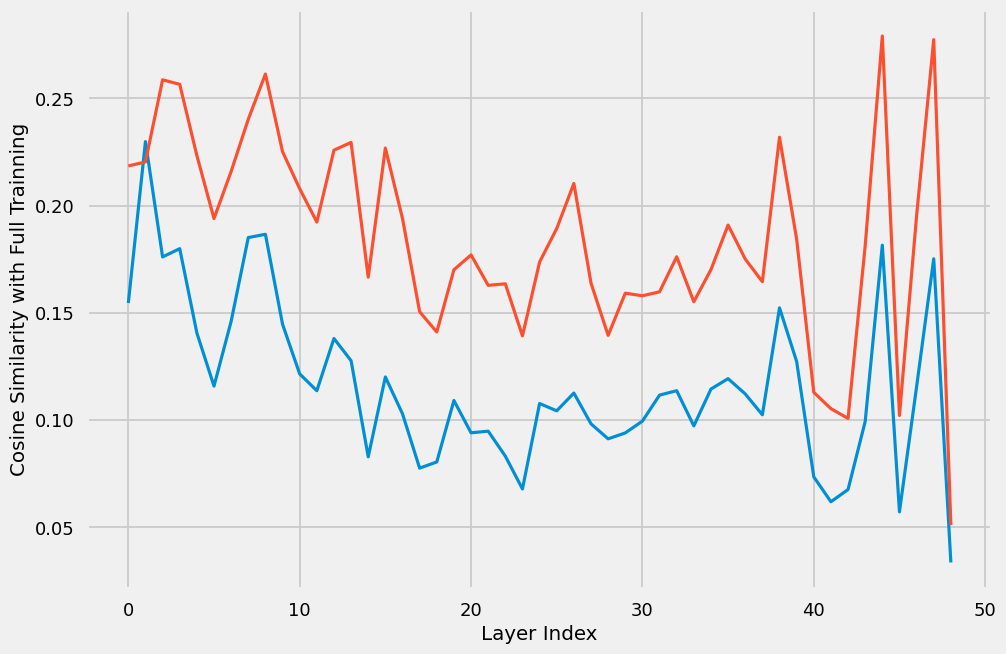}
        \caption{part-3}
        \label{fig:part3}
    \end{subfigure}
    \begin{subfigure}{0.47\linewidth}
        \centering
        \includegraphics[width=\linewidth]{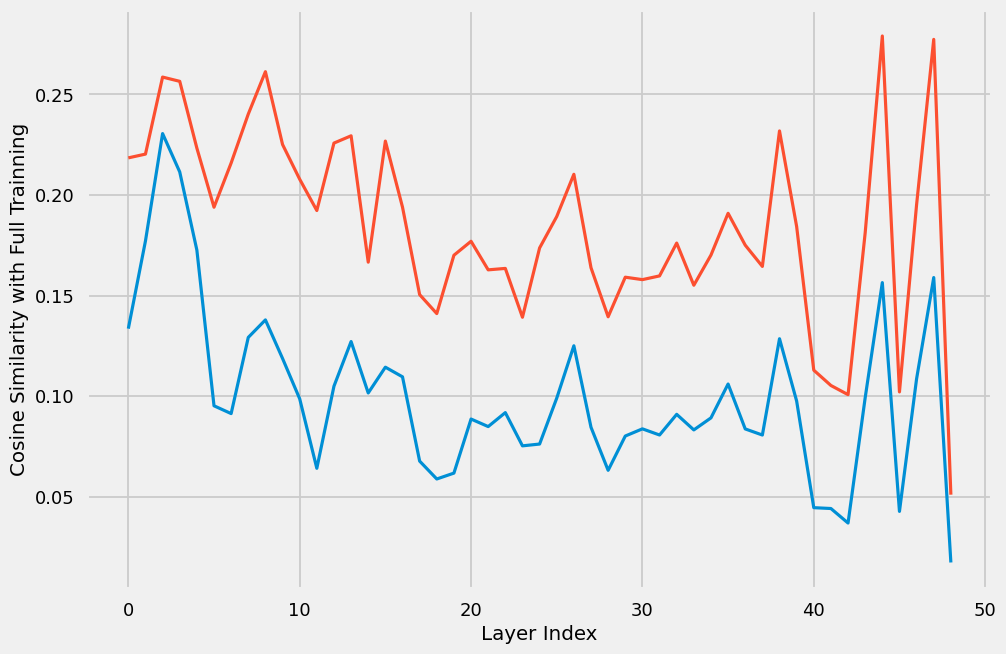}
        \caption{part-4}
        \label{fig:part4}
    \end{subfigure}
    \caption{An explanation of the direction in which the data vector is moving}
    \label{fig:4part}
\end{figure}

To further justify the rationality of our method, we calculated the data vectors using models trained on the full dataset (Full Training), models trained on four partial datasets, and the hybrid model, respectively, as well as the pre-trained model. Since the base model used was ResNet50, we only considered the data vectors calculated from the weights of the convolutional layers. We computed the cosine similarity between the data vectors of the models trained on the four partial datasets and the hybrid model, respectively, and the data vector of the model trained on the full dataset, layer by layer. The visualization results of the cosine similarity by layer are shown in \autoref{fig:4part}. In the figure, the red line represents the cosine similarity between the data vector of the hybrid model and the data vector of the model trained on the full dataset, while the blue line represents the cosine similarity between the data vectors of the models trained on different partial datasets and the data vector of the model trained on the full dataset. It can be seen that the cosine similarity between the data vector of the hybrid model and the data vector of the full dataset is significantly improved in comparison, which also proves that the hybrid model can make the parameters closer to the effect of the parameters trained on the full dataset.

% \printbibliography
% \bibliographystyle{unsrt}  
% \bibliography{ref}  

% \newpage
\appendix

\section{Appendix}
\label{appendix}

\subsection{Endoscopic Bladder Tissue}
This dataset is derived from the article 
 \cite{lazo2023semi} and comprises a total of 1754 endoscopic images from 23 patients who underwent Transurethral Resection of Bladder Tumor (TURBT). The images in this dataset are annotated based on histopathological analysis of the excised tissues. Typically, endoscopic imaging employs White Light Imaging (WLI) technology, and if conditions permit, Narrow Band Imaging (NBI) technology is used. According to the classification standards of the World Health Organization (WHO) and the International Society of Urological Pathology, the bladder is categorized into four distinct classes, including two cancerous tissue categories: Low-Grade Cancer (LGC) and High-Grade Cancer (HGC), as well as two non-tumor lesion categories, namely Cystitis (NTL) caused by infection or other inflammatory factors and Non-Suspicious Tissue (NST). This dataset records the name of each frame, imaging type (NBI or WLI), tissue type (HGC, LGC, NTL, NST), and the dataset used for training the classifier (train/val/test).

\subsection{PAD-UFES-20}
The PAD-UFES-20 \cite{pacheco2020pad} dataset comprises 2298 dermatological images captured by various smartphones from 1373 patients. This dataset covers six categories, including three types of skin cancer: Basal Cell Carcinoma (BCC), Squamous Cell Carcinoma (SCC), and Melanoma (MEL); as well as three types of skin diseases: Actinic Keratosis (ACK), Seborrheic Keratosis (SEK), and Nevus (NEV). Notably, the SCC category also includes Bowen's Disease (BOD). All instances of BCC, SCC, and MEL in the dataset have been verified through biopsy, while the remaining categories have been jointly diagnosed by a team of dermatologists. For each image in the dataset, the authors provide information such as image category labels, patient age, location of skin lesions, and diameter of skin lesions.

\subsection{Retina}
The Retina \cite{Retina} dataset comprises 601 retinal fundus images, covering four categories: 300 normal images, 100 cataract images, 101 glaucoma images, and 100 retina disease images. Each year, a large number of people worldwide are diagnosed with visual impairments, with millions suffering from retinal diseases alone. As the global population ages, this group of visually impaired individuals is expected to grow significantly, with glaucoma, cataract, and retinal diseases being among the most common conditions. In recent years, artificial intelligence technologies, represented by deep learning algorithms, have been actively driving advancements in the field of medical imaging. The authors hope that the Retina dataset will assist researchers in developing vision impairment disease recognition models and software based on retinal fundus images, enabling doctors to perform rapid and accurate disease diagnosis and treatment efficacy analysis.

\end{document}